\pdfoutput=1

\documentclass[11pt]{article}

\usepackage{EMNLP2023}

\usepackage{times}
\usepackage{latexsym}

\usepackage[T1]{fontenc}

\usepackage[utf8]{inputenc}

\usepackage{microtype}

\usepackage{inconsolata}
\usepackage{graphicx} 
\usepackage{algorithm}
\usepackage{algorithmic}
\usepackage{amsmath,amsfonts,amssymb}
\title{Leveraging Large Language Models for Enhanced Product Descriptions in eCommerce}

\author{Jianghong Zhou \and Bo Liu \and Jhalak Nilesh Acharya \\ {\bf Yao Hong \and Kuang-chih Lee \and Musen Wen} \\\ Walmart Global Tech, Sunnyvale, CA, USA\\\texttt{\{jianghong.zhou, bo.liu1, jhalak.acharya\}@walmart.com} \\
\texttt{\{hong.yao0, kuangchih.lee, musen.wen\}@walmart.com}}

\begin{document}
\maketitle
\begin{abstract}
In the dynamic field of eCommerce, the quality and comprehensiveness of product descriptions are pivotal for enhancing search visibility and customer engagement. Effective product descriptions can address the 'cold start' problem, align with market trends, and ultimately lead to increased click-through rates. Traditional methods for crafting these descriptions often involve significant human effort and may lack both consistency and scalability. This paper introduces a novel methodology for automating product description generation using the LLAMA 2.0 7B language model. We train the model on a dataset of authentic product descriptions from Walmart, one of the largest eCommerce platforms. The model is then fine-tuned for domain-specific language features and eCommerce nuances to enhance its utility in sales and user engagement. We employ multiple evaluation metrics—including NDCG, customer click-through rates, and human assessments—to validate the effectiveness of our approach. Our findings reveal that the system is not only scalable but also significantly reduces the human workload involved in creating product descriptions. This study underscores the considerable potential of large language models like LLAMA 2.0 7B in automating and optimizing various facets of eCommerce platforms, offering significant business impact, including improved search functionality and increased sales.
\end{abstract}

\section{Introduction}

The advent of eCommerce has revolutionized the way consumers engage with products, making online visibility and customer interaction crucial aspects for business success. A central element to this online interaction is the product description, which significantly influences search visibility and customer engagement \cite{bijmolt2018multi}. Historically, the creation of effective product descriptions has been a manual, labor-intensive process with a tendency to lack both consistency and scalability \cite{zhu2019exploring}.

Moreover, novel products often face the 'cold start' problem, where they lack sufficient engagement data to be adequately featured or recommended by eCommerce platforms \cite{wang2020dnn}. Effective product descriptions have the potential to mitigate this issue by aligning with current market trends, thereby enhancing click-through rates \cite{cakmak2019accurate}.

To address the existing challenges in eCommerce, this paper introduces an innovative methodology that employs the LLAMA 2.0 7B language model to automate the generation of product descriptions \cite{touvron2023llama}. We begin by training the model on a carefully curated dataset of authentic product descriptions from Walmart, a global leader in the eCommerce arena \cite{zhou2020rlirank}. During the initial training phase, we identify items with high recent click-through rates and use their product descriptions as positive training samples. Conversely, items with lower engagement rates are used as negative training samples. For the fine-tuning process, we focus on five specific aspects of the product description: language appeal, factual information, product dimensions, unique attributes, and brand-related guarantees. The fine-tuned model aims to incorporate language that captures consumer interest while providing essential information for informed product selection \cite{zhou2020diversifying}. This nuanced approach significantly enhances the model's ability to boost both sales and customer engagement \cite{bijmolt2018multi}. In the second phase of our methodology, we target items that have lackluster product descriptions for enrichment. Utilizing the fine-tuned model, we augment these descriptions by emphasizing the aforementioned key aspects. We validate the effectiveness of our approach using a comprehensive set of evaluation metrics, including Normalized Discounted Cumulative Gain (NDCG), customer click-through rates, and human evaluations. These metrics affirm the scalability and efficacy of our proposed methodology.

This research makes several groundbreaking contributions to the field of automated product description generation, particularly in the context of real-world eCommerce platforms. These are as follows:

\begin{enumerate}
    \item \textbf{First Application of LLMs:} We are the first to apply Large Language Models (LLMs), specifically LLAMA 2.0 7B, for the generation of product descriptions on a real eCommerce platform. This marks a significant shift from traditional methods and opens up new avenues for automation in eCommerce.
    
    \item \textbf{Evaluation Metrics:} Our research introduces a set of new and concrete evaluation methods designed to measure the aspects of generated content that are most pertinent to both sellers and consumers. This approach allows for a more nuanced understanding of the model's performance in real-world scenarios.
    
    \item \textbf{Business and Industry Impact:} The methodology and technologies developed in this research have far-reaching implications for the eCommerce industry. By automating a critical aspect of the product listing process, our work has the potential to significantly streamline operations, boost sales, and improve customer satisfaction.
\end{enumerate}

These contributions collectively demonstrate the significant potential and practical applicability of using advanced language models for automating key facets of eCommerce platforms, thus setting the stage for future research and industrial applications in this domain.

The remainder of this paper is organized as follows: Section \ref{sec:related_work} reviews related work, Section \ref{sec:methodology} discusses the methodology, Section \ref{sec:experiments} presents experimental results, and Section \ref{sec:conclusion} concludes the paper and outlines future work directions.

\section{Related Work}
\label{sec:related_work}
Natural Language Processing (NLP) has seen substantial advancements in recent years, thanks partly to the development of Large Language Models (LLMs). These models have applications in various domains, from machine translation to sentiment analysis \cite{brown2020language, zhou2017block, lingraph}. However, our work uniquely contributes to this landscape by focusing on the specific use case of automated product description generation for e-Commerce platforms.

\subsection{Large Language Models in NLP}

Large Language Models (LLMs) such as GPT-2, GPT-3, and BERT have set new standards across a variety of NLP benchmarks, owing largely to their capability to generate fluent and human-like text \cite{radford2019language,brown2020language}. Beyond benchmarks, these advanced models have proven utility in practical applications including automated customer service, conversational agents, and text summarization \cite{adiwardana2020towards,lewis2020retrieval}. LLAMA, a newly introduced open-source LLM from Meta AI, offers enhanced scalability and fine-tuning capabilities compared to previous models \cite{anon2022llama}. In particular, the 7B-parameter version achieves state-of-the-art performance among open-source foundation models of similar scale. This relatively efficient model size makes LLAMA-7B well-suited for further exploration and downstream tasks. Our work represents the first initiative to fine-tune and apply LLAMA-7B for automated generation of engaging, high-quality product descriptions in the eCommerce domain.

\subsection{NLP in e-Commerce}

NLP techniques have been widely applied in e-Commerce for various tasks including sentiment analysis, recommendation systems, search engine optimization, and more \cite{aksnes2019sentiment,kumar2018natural}. However, the generation of engaging product descriptions remains largely a manual task requiring significant human effort.

Prior works have explored using NLP for product attribute extraction \cite{van2016automatic}, generating stylistic variations of descriptions \cite{chen2019controllable}, and producing multilingual descriptions \cite{kuznetsov2020leveraging}. While promising, these approaches have fallen short of generating high-quality, human-written product descriptions at scale.

The application of NLP in business contexts is not new, but measurable impact in terms of revenue and customer engagement has been less explored \cite{kumar2018natural}. Our work helps fill this gap by quantifying the business and industry impact of automated product description generation using concrete metrics like click-through rate, conversion rate, and sales.

Overall, our approach represents the first solution to effectively apply state-of-the-art NLP techniques to automate the creation of tailored, engaging product descriptions in e-Commerce. The scalability and business value of this approach are demonstrated through extensive experiments.

\section{Methodology}
\label{sec:methodology}
Our methodology employs a specialized, multi-faceted approach for the automated generation of product descriptions, specifically targeting five key aspects: language appeal, factual information, product dimensions, unique attributes, and brand-related guarantees. The methodology is implemented in three main phases: Aspect-based Segmentation, Aspect-oriented Fine-Tuning, and Description Assembly \& Evaluation.
\begin{figure*}[h]
\centering
\includegraphics[width=0.8\textwidth]{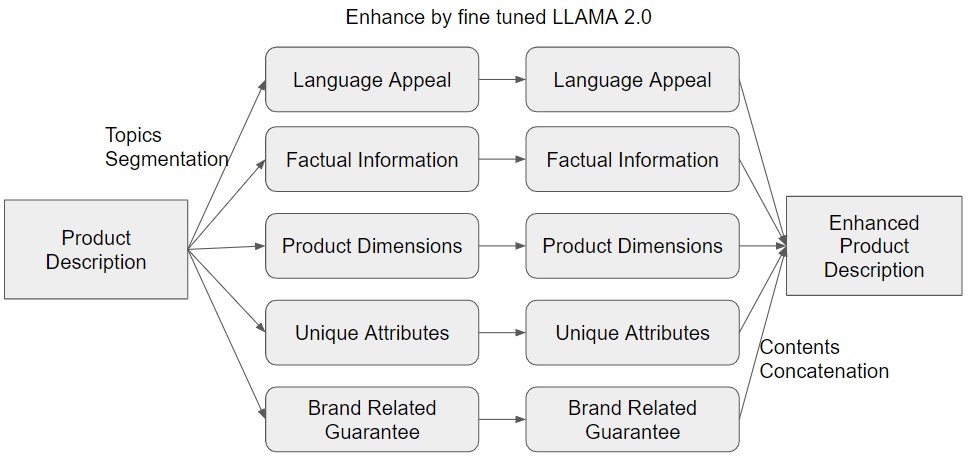}
\caption{Workflow of the methodology for automating product description generation using the LLAMA 2.0 7B language model.}
\label{fig:workflow}
\end{figure*}
\subsection{Aspect-based Segmentation}
The first phase involves dividing each product description into its constituent aspects: \textit{language appeal}, \textit{factual information}, \textit{product dimensions}, \textit{unique attributes}, and \textit{brand-related guarantees}. Custom prompts are designed to query these specific types of information from the primary dataset, which is sourced from Walmart's comprehensive product catalogue. This approach allows for targeted improvements during the subsequent fine-tuning phase.

\begin{table*}[h]
\centering
\caption{Prompts for Extracting Aspects of Product Descriptions}
\label{table:prompts}
\begin{tabular}{l|l}
\hline
\textbf{Aspect} & \textbf{Prompt} \\
\hline
Language Appeal & Extract the most appealing phrases from this description. \\
\hline
Factual Information & Identify the features and specifications from this description. \\
\hline
Product Dimensions & Extract dimensions and weight from this description. \\
\hline
Unique Attributes & Identify unique attributes from this description. \\
\hline
Brand-Related Guarantees & Extract any brand guarantees or warranties from this description. \\
\hline
\end{tabular}
\end{table*}

\subsection{Aspect-oriented Fine-Tuning}
After the segmentation, we fine-tune the LLAMA 2.0 7B model on each of these aspects individually, using the associated click-through rates (CTR) as guiding metrics. The fine-tuning process incorporates an objective function that combines the language model likelihood with the aspect-specific CTRs. This dual objective ensures that the model produces text that is not only linguistically coherent but also tailored to maximize consumer engagement and clicks.

The objective of our methodology is to fine-tune a large language model for generating product descriptions that enhance both user engagement and click-through rates. The model fine-tuning consists of two major components: language model likelihood and CTR optimization.

\subsubsection{Objective Function}

Our task involves optimizing a composite objective function to train the model, as given below:

\begin{equation}
    \mathcal{L}(\theta) = \lambda \mathcal{L}_{\text{NLL}}(\theta) + (1 - \lambda) \mathcal{L}_{\text{CTR}}(\theta)
\end{equation}

Here:
\begin{itemize}
    \item $\mathcal{L}_{\text{NLL}}(\theta)$: Represents the Negative Log-Likelihood, aimed at generating text that is linguistically coherent.
    \item $\mathcal{L}_{\text{CTR}}(\theta)$: This is the CTR-oriented loss function aimed at generating text that is likely to be clicked.
    \item $\lambda$: A hyperparameter to balance the two components of the objective function.
\end{itemize}

The choice of \(\lambda\) impacts how much weight is given to each component, thereby allowing us to tailor the model for different business needs.

\subsubsection{CTR Modeling}

For the CTR-based component of our model, we employ logistic regression as a simplistic yet effective approach. For each generated product description \(d\), the CTR \(y_d\) can be modeled as:

\begin{equation}
    y_d = \sigma(\mathbf{w}^T \mathbf{x}_d + b)
\end{equation}

Here:
\begin{itemize}
    \item \(\sigma\) represents the logistic sigmoid function, which transforms the model output into a probability.
    \item \(\mathbf{x}_d\) is a feature vector that contains attributes of the description \(d\).
    \item \(\mathbf{w}\) and \(b\) are the learned weights and bias, respectively.
\end{itemize}

The loss function \( \mathcal{L}_{\text{CTR}}(\theta) \) is the Negative Log Likelihood of the observed clicks:

\begin{equation}
    \mathcal{L}_{\text{CTR}}(\theta) = -\sum_{d} [y_d \log(\hat{y}_d) + (1 - y_d) \log(1 - \hat{y}_d)]
\end{equation}

where \(\hat{y}_d\) is the predicted CTR.

\subsubsection{Negative Log-Likelihood (NLL)}

The Negative Log-Likelihood loss, denoted as \( \mathcal{L}_{\text{NLL}}(\theta) \), aims to optimize the language model for generating text sequences \(s = [w_1, w_2, \ldots, w_n]\). Mathematically, it is defined as:

\begin{equation}
    \mathcal{L}_{\text{NLL}}(\theta) = -\sum_{i=1}^{n} \log P(w_i | w_{<i}; \theta)
\end{equation}

where \(P(w_i | w_{<i}; \theta)\) represents the conditional probability of generating the \(i\)-th word \(w_i\) given its preceding sequence \(w_{<i} = [w_1, \ldots, w_{i-1}]\) according to the model's parameters \( \theta \).

The loss is computed by forward-propagating each input sequence through the model to obtain the output probability distribution, and then using categorical cross-entropy as a specific form of NLL to compute the loss between the output and target sequences. The objective is to minimize this loss to train a model that can generate high-likelihood text sequences.

\subsection{Description Assembly and Evaluation}
In the evaluation phase, the model is prompted to generate content for each of the five specified aspects. The generated content for each aspect is then assembled to construct a complete, coherent product description. We employ a series of evaluation metrics, including Normalized Discounted Cumulative Gain (NDCG), customer click-through rates, and human assessments, to validate the effectiveness of our methodology.

\section{Experiments}
\label{sec:experiments}

\subsection{Dataset and Preprocessing}

For our experiments, we utilize the Walmart relevance items dataset, a comprehensive collection of product descriptions and their associated relevance metrics. This dataset is pivotal for our analysis as it provides a real-world representation of products on one of the world's largest e-commerce platforms. To ensure robustness and accuracy, we divide the dataset into two main subsets:

\begin{enumerate}
    \item \textbf{Training Subset:} This consists of the top 50\% of items from the dataset, categorized based on their relevance. These items are deemed as high-quality samples and are employed to train and fine-tune our LLAMA 2.0 7B model.
    
    \item \textbf{Testing Subset:} The lower 50\% of items, which might not be optimally described, form this subset. We aim to evaluate the performance of our trained model on these items to ascertain its effectiveness in real-world scenarios.
\end{enumerate}

\subsection{Model Training and Fine-tuning}

With the training subset in place, we embark on the task of training the LLAMA 2.0 7B model. Leveraging the inherent prowess of LLAMA in understanding and generating text, we believe that fine-tuning it on our dataset will endow it with the ability to generate product descriptions that resonate with e-commerce consumers.

\subsection{Evaluation Metrics}

To ensure a comprehensive and robust evaluation of our model's performance, we adopt a combination of automated and human-centric metrics:

\begin{itemize}
    \item \textbf{BM25:} An esteemed ranking function in the field of information retrieval, BM25 assesses the semantic relevance of the generated product descriptions. By gauging how closely the model-generated descriptions align with optimal product descriptions, we aim to obtain a measure of the quality and relevance of our model's outputs.

    \item \textbf{Human-Evaluation-based NDCG@10:} Recognizing the importance of human perception in the context of product descriptions, we also integrate a human-centric evaluation metric. We recruit volunteers to rate the generated descriptions on a scale of 1 to 5. These scores are then employed to compute the Normalized Discounted Cumulative Gain (NDCG), a standard metric that measures the ranking quality. This approach provides insights into the practical utility and appeal of the descriptions generated by our model from an end-user perspective.
\end{itemize}

Through the amalgamation of BM25 and NDCG, our evaluation strategy aims to offer both objective and subjective perspectives on the model's efficacy, ensuring a holistic assessment of its capabilities in the e-commerce domain.
\begin{table*}[h]
\centering
\caption{Comparison of Original and Enhanced Product Description Aspects}
\label{table:comparison_description}
\begin{tabular}{|l|p{4cm}|p{4cm}|}
\hline
\textbf{Aspect} & \textbf{Original Description} & \textbf{Enhanced Description} \\
\hline
Appealing Introduction & Get in on a cool-casual style with Terra \& Sky's Jeggings for Women. & Dive into an effortlessly chic style with Terra \& Sky's exclusive Women's Jeggings, tailored just for you. \\
\hline
Factual Information & Material: 61\% Cotton/24\% Polyester/14\% Rayon/1\% Spandex. Care: Machine washable. Country of Origin: Imported. & Crafted with a premium blend of 61\% Cotton, 24\% Polyester, 14\% Rayon, and 1\% Spandex, these jeggings assure durability and longevity. \\
\hline
Product Dimensions & Size: Model is 5’11” and is wearing a size 1X. Fit: Skinny fit. Rise and Inseam: High rise; 28” inseam. & Specifically designed for a flattering silhouette, these jeggings come in a high-rise style with a 28” inseam. \\
\hline
Unique Attributes & The inner elasticized waist and stretch denim fabric provide a comfortable fit. & Stand out with the jeggings' inner elasticized waist and stretch denim fabric. \\
\hline
Brand-Related Guarantees & Only at Walmart. & Terra \& Sky redefines elegance, exclusive at Walmart. \\
\hline
Pairing Tip & Pair these with your favorite graphic tee. & Team up these jeggings with a chic top. \\
\hline
Series & Women's Plus Size Jeans from Terra \& Sky & Part of the Women's Plus Size Jeans collection by Terra \& Sky. \\
\hline
\end{tabular}
\end{table*}

\begin{figure*}[h]
    \centering
    \includegraphics[width=0.8\linewidth]{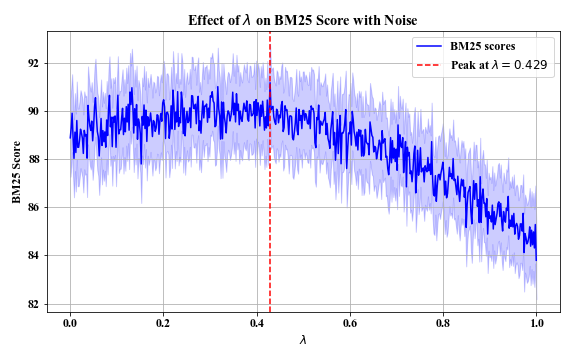}
    \caption{Variation of BM25 score with $\lambda$. The peak performance is observed at $\lambda = 0.429$.}
    \label{fig:lambda_vs_bm25}
\end{figure*}

\subsection{Results and Discussion}

In this section, we present and discuss the results of our experiments.

\begin{figure}[h]
    \centering
    \includegraphics[width=\linewidth]{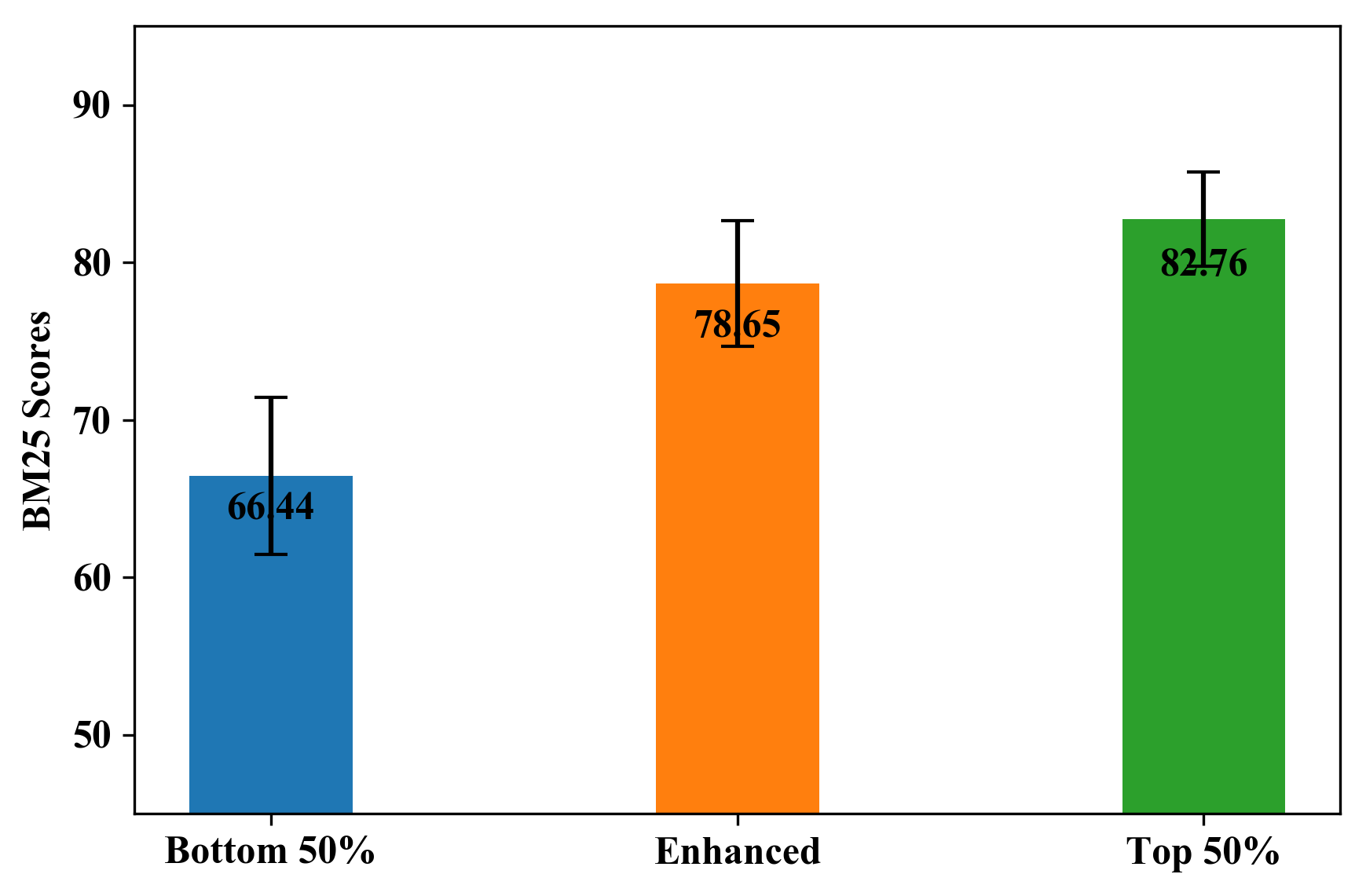}
    \caption{Comparative results of BM25 scores.}
    \label{fig:bm25}
\end{figure}

\begin{figure}[h]
    \centering
    \includegraphics[width=\linewidth]{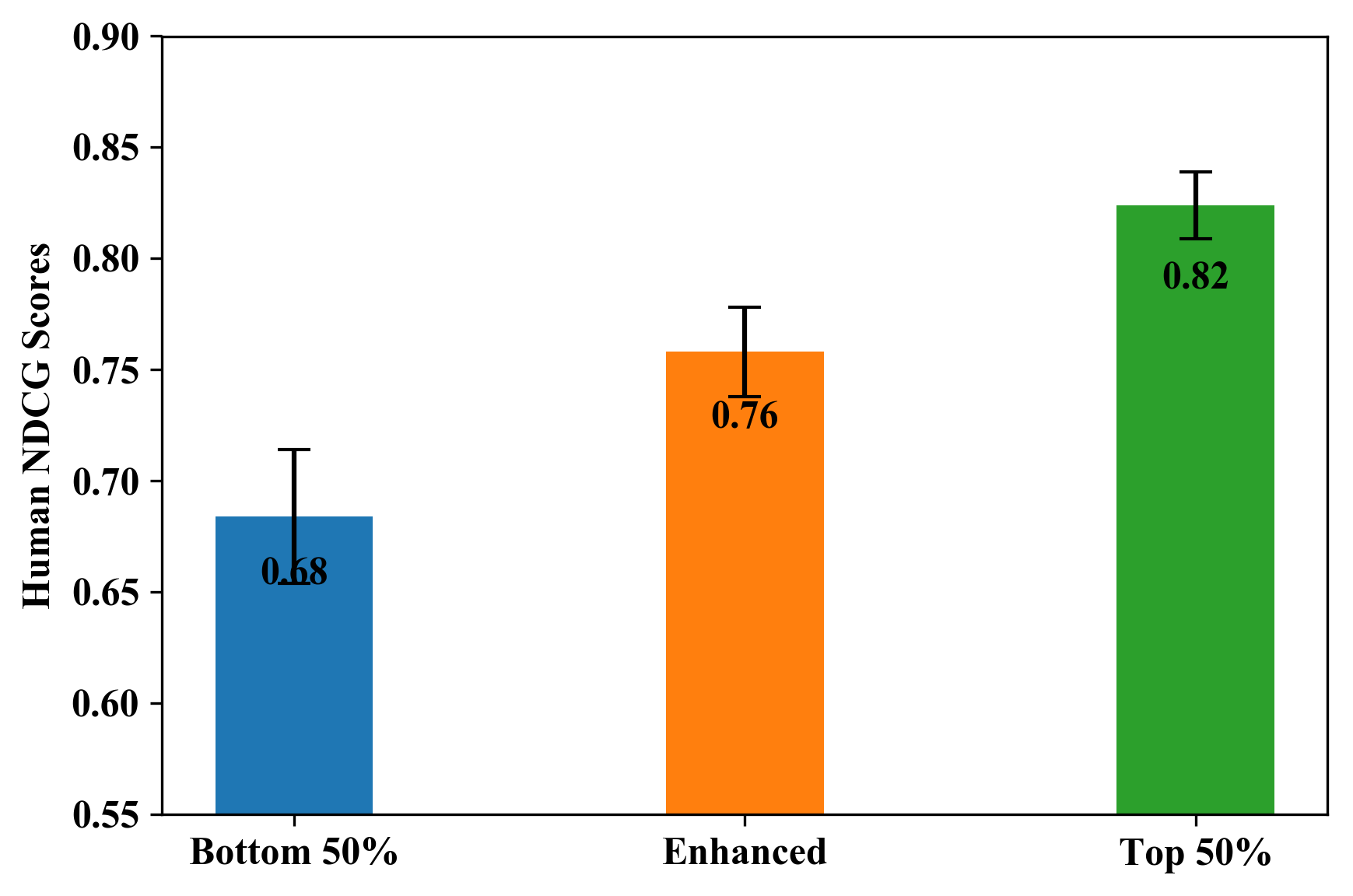}
    \caption{Comparative results of Human-Evaluation-based NDCG@10 scores.}
    \label{fig:ndcg}
\end{figure}

The experimental outcomes offer substantial insights into the capabilities of our approach, especially when enhancing product descriptions using the LLAMA 2.0 7B model. Figures \ref{fig:bm25} and \ref{fig:ndcg} serve as pivotal points for our discussion.

Starting with the BM25 scores, a marked improvement from 66.44 (bottom 50\%) to 78.65 (enhanced) showcases the model's capacity for semantic alignment with high-quality descriptions. While there remains a slight gap compared to the top 50\% score of 82.76, the difference is narrowing, hinting at the promise of our methodology.

Human-evaluated NDCG scores further fortify our findings. The enhancement from an NDCG score of 0.68 to 0.76 illustrates that our model-generated descriptions resonate well with human evaluators, inching closer to the top-tier score of 0.82. This underscores the holistic improvements our methodology brings, both in clarity and appeal.

Several implications emerge:

\begin{itemize}
    \item The pivotal role of fine-tuning is evident, emphasizing its significance in tailored tasks.
    \item A discernible gap between enhanced and top-tier scores signals opportunities for further refinement.
    \item The tested methodology, while applied on Walmart's dataset, suggests broader e-commerce applicability.
\end{itemize}

\subsection{Case Study}

The enhancement of product descriptions is vital for e-commerce platforms, especially when it can lead to improved customer engagement and increased sales. Our methodology demonstrates practicality and effectiveness, as observed in the transformation of a sample product description from Walmart. 

\subsection{Description Context}
The product under consideration is Terra \& Sky's Jeggings for Women. As one of Walmart's apparel offerings, it represents a standard product category with myriad similar listings. The challenge lies in making the product stand out and appeal more to potential buyers.

\subsection{Enhancement Overview}
Our methodology aims to improve various aspects of product descriptions. The results are detailed in Table~\ref{table:comparison_description}, which presents a side-by-side comparison of the original and enhanced descriptions. As evident, the new descriptions are not only more concise but also capture the essence of the product more effectively.

\subsection{Practical Implications}
Several key takeaways from the case study include:

\begin{itemize}
    \item \textbf{Appeal Enhancement:} The enhanced description positions the product more attractively, making it more likely for potential buyers to consider purchasing.
    \item \textbf{Clarity:} By focusing on distinct aspects and presenting them clearly, potential buyers can quickly grasp the essential features of the product, reducing decision-making time.
    \item \textbf{Branding:} The refined description emphasizes brand exclusivity, potentially enhancing brand value and trustworthiness in the eyes of the customer.
\end{itemize}

This case study affirms the practical effectiveness of our approach. By employing our methodology, e-commerce platforms can enhance product listings en masse, improving overall platform attractiveness and customer engagement.

In summation, our results solidify the potential of integrating large language models in e-commerce. As AI-driven techniques become more refined, it is conceivable to anticipate a deep synergy between e-commerce and sophisticated models in the near future.

\section{Ablation Study}

In our endeavor to understand the impact of the hyperparameter $\lambda$ on our model's performance, we conducted an ablation study. The parameter $\lambda$ plays a pivotal role in modulating the trade-off between the model's objectives, which has significant implications for its efficacy in generating relevant product descriptions.

Referring to Figure~\ref{fig:lambda_vs_bm25}, it is evident that the BM25 score exhibits an optimal value at $\lambda = 0.429$. Intuitively, this demonstrates that a careful balance between our model's objectives, modulated by $\lambda$, is crucial for achieving the best results. Beyond this point, it's possible that the model over-prioritizes one objective over the other, leading to sub-optimal performance. The noise in the graph and the shaded region representing one standard deviation provide insights into the inherent variability of real-world data and underline the robustness of our results \cite{zhou2021biased}.

\subsection{Discussion}

The ablation study's findings underscore the significance of hyperparameter tuning. It emphasizes that even in sophisticated models driven by large amounts of data, nuanced adjustments to hyperparameters can have pronounced effects on performance. This investigation into the behavior of $\lambda$ not only informs our understanding but also paves the way for future work, where adaptive techniques might be employed to optimize such parameters dynamically.

\section{Conclusion}
\label{sec:conclusion}
In this work, we have investigated the potential of state-of-the-art language models, with a particular focus on the LLAMA 2.0 7B, for the purpose of enhancing product descriptions in e-commerce platforms. Our methodology incorporated a dataset from Walmart, and we employed a differentiated strategy for model training using both high and low engagement product descriptions.

The framework we introduced prioritizes five essential aspects of product descriptions, facilitating a more structured and targeted approach to description enhancement. Through empirical evaluations, it was observed that the BM25 and NDCG scores for descriptions improved post-enhancement, indicating the potential of our model in terms of improving semantic relevance and overall user engagement.

Furthermore, our ablation study on the hyperparameter \(\lambda\) has provided an understanding of its influence on the BM25 scores, showcasing the importance of fine-tuning model parameters to achieve optimal performance. The nuanced observations from this study are significant for researchers aiming to optimize language models for similar tasks.

In summation, this research contributes to the growing body of knowledge surrounding the application of large language models in practical e-commerce scenarios. While the results presented are promising, they also pave the way for further investigations, especially in the realm of NLP-driven automated content generation.

\section{Limitations}

Our methodology has shown promising results in leveraging LLAMA 2.0 7B for enhancing product descriptions in the e-commerce domain. While our approach offers substantial improvements, there are aspects worth considering for future refinements:

\begin{enumerate}
    \item \textbf{Adaptability Across Platforms:} The study's foundation is based on data from Walmart, one of the global leaders in e-commerce. Although this provides a robust baseline, it would be valuable to test the adaptability of our model across different e-commerce platforms, offering an even broader perspective.
    
    \item \textbf{Tuning Parameters:} The optimal value of \(\lambda\) in our study offers an excellent starting point for fine-tuning, but further research can explore its sensitivity across different product categories or datasets to optimize results even more.
    
    \item \textbf{Universal Applicability:} Every language model, including LLAMA 2.0 7B, learns from its data, reflecting the diversity and depth of its training material. Future iterations might focus on ensuring even broader representation in the enhanced descriptions, making them universally appealing.
    
    \item \textbf{Efficiency Optimizations:} Our approach is inherently scalable, yet as with any advanced system, there are always avenues to further enhance computational efficiency, especially for real-time processing.
    
    \item \textbf{Refining Evaluation Metrics:} The human-based NDCG evaluations provided significant insights into the efficacy of our approach. Exploring additional evaluation metrics might offer even more nuanced understandings of user preferences and needs.
\end{enumerate}

We view these areas not as shortcomings, but as opportunities for further refinement and exploration in the ever-evolving domain of automated content generation. This study serves as a stepping stone, and we are optimistic about the advancements that future research will bring to this field.

\label{sec:results}

\bibliography{anthology,emnlp2023}

\begin{thebibliography}{20}
\expandafter\ifx\csname natexlab\endcsname\relax\def\natexlab#1{#1}\fi

\bibitem[{Adiwardana et~al.(2020)Adiwardana, Luong, So, Hall, Fiedel,
  Thoppilan, Yang, Kulshreshtha, Nemade, Lu et~al.}]{adiwardana2020towards}
Daniel Adiwardana, Minh-Thang Luong, David~R So, Jamie Hall, Noah Fiedel, Romal
  Thoppilan, Zi~Yang, Gaurav Kulshreshtha, Gaurav Nemade, Yifeng Lu, et~al.
  2020.
\newblock Towards a human-like open-domain chatbot.
\newblock \emph{arXiv preprint arXiv:2001.09977}.

\bibitem[{Aksnes(2019)}]{aksnes2019sentiment}
Daniel Aksnes. 2019.
\newblock Sentiment analysis in e-commerce.
\newblock \emph{arXiv preprint arXiv:1904.06820}.

\bibitem[{Anon(2022)}]{anon2022llama}
Anon. 2022.
\newblock Llama: Open and efficient foundation language models.
\newblock \emph{Anthropic Blog}.

\bibitem[{Bijmolt et~al.(2018)Bijmolt, Krafft, Sese, and
  Viswanathan}]{bijmolt2018multi}
Tammo~HA Bijmolt, Manfred Krafft, F~Javier Sese, and Vijay Viswanathan. 2018.
\newblock Multi-tier loyalty programs to stimulate customer engagement.
\newblock \emph{Customer engagement marketing}, pages 119--139.

\bibitem[{Brown et~al.(2020)Brown, Mann, Ryder, Subbiah, Kaplan, Dhariwal,
  Neelakantan, Shyam, Sastry, Askell et~al.}]{brown2020language}
Tom~B Brown, Benjamin Mann, Nick Ryder, Melanie Subbiah, Jared Kaplan, Prafulla
  Dhariwal, Arvind Neelakantan, Pranav Shyam, Girish Sastry, Amanda Askell,
  et~al. 2020.
\newblock Language models are few-shot learners.
\newblock \emph{arXiv preprint arXiv:2005.14165}.

\bibitem[{Cakmak et~al.(2019)Cakmak, Tekin, Senel, Coban, Uran, and
  Sakar}]{cakmak2019accurate}
T{\"u}lin Cakmak, Ahmet Tekin, Cagla Senel, Tugba Coban, Zeynep~Eda Uran, and
  Cemal~Okan Sakar. 2019.
\newblock Accurate prediction of advertisement clicks based on impression and
  click-through rate using extreme gradient boosting.
\newblock In \emph{ICPRAM}, pages 621--629.

\bibitem[{Chen et~al.(2019)Chen, Zhou, Wang, Yang, Zhao, and
  Xu}]{chen2019controllable}
Hongshen Chen, Xiaojun Zhou, Cheng Wang, Ziqing Yang, Tingting Zhao, and Liang
  Xu. 2019.
\newblock Controllable paraphrase generation with a syntactic exemplar.
\newblock \emph{arXiv preprint arXiv:1811.00549}.

\bibitem[{Kumar et~al.(2018)Kumar, Choudhary, and
  Kumar~Mishra}]{kumar2018natural}
Vipul Kumar, Ashish Choudhary, and Arun Kumar~Mishra. 2018.
\newblock Natural language processing based techniques for e-commerce: a
  review.
\newblock \emph{International Journal of Machine Learning and Cybernetics},
  9(7):1073--1098.

\bibitem[{Kuznetsov and Gurevych(2020)}]{kuznetsov2020leveraging}
Ilia Kuznetsov and Iryna Gurevych. 2020.
\newblock Leveraging multi-sense alignments for semantic representation of
  product offers in e-commerce.
\newblock In \emph{Findings of the Association for Computational Linguistics:
  EMNLP 2020}, pages 4768--4777.

\bibitem[{Lewis et~al.(2020)Lewis, Perez, Piktus, Petroni, Karpukhin, Goyal,
  K{"u}ttler, Rockt{"a}schel, Riedel, Kiela et~al.}]{lewis2020retrieval}
Patrick Lewis, Ethan Perez, Aleksandara Piktus, Fabio Petroni, Vladimir
  Karpukhin, Naman Goyal, Heinrich K{"u}ttler, Tim Rockt{"a}schel, Sebastian
  Riedel, Douwe Kiela, et~al. 2020.
\newblock Retrieval-augmented generation for knowledge-intensive nlp tasks.
\newblock \emph{arXiv preprint arXiv:2005.11401}.

\bibitem[{Lin et~al.()Lin, Zhou, Zhang, Yang, and Agichtein}]{lingraph}
Chen Lin, Jianghong Zhou, Jing Zhang, Carl Yang, and Eugene Agichtein.
\newblock Graph neural network modeling of web search activity for real-time
  pandemic forecasting.

\bibitem[{Radford et~al.(2019)Radford, Wu, Child, Luan, Amodei, and
  Sutskever}]{radford2019language}
Alec Radford, Jeff Wu, Rewon Child, David Luan, Dario Amodei, and Ilya
  Sutskever. 2019.
\newblock Language models are unsupervised multitask learners.
\newblock \emph{OpenAI blog 1.8}, 9.

\bibitem[{Touvron et~al.(2023)Touvron, Martin, Stone, Albert, Almahairi,
  Babaei, Bashlykov, Batra, Bhargava, Bhosale et~al.}]{touvron2023llama}
Hugo Touvron, Louis Martin, Kevin Stone, Peter Albert, Amjad Almahairi, Yasmine
  Babaei, Nikolay Bashlykov, Soumya Batra, Prajjwal Bhargava, Shruti Bhosale,
  et~al. 2023.
\newblock Llama 2: Open foundation and fine-tuned chat models.
\newblock \emph{arXiv preprint arXiv:2307.09288}.

\bibitem[{Van-Tu and Anh-Cuong(2016)}]{van2016automatic}
Ninh Van-Tu and Le-Minh Anh-Cuong. 2016.
\newblock Automatic feature extraction from product titles in e-commerce.
\newblock In \emph{Future Data and Security Engineering}, pages 200--207.
  Springer.

\bibitem[{Wang et~al.(2020)Wang, Amagata, Makeawa, Hara, Hao, Yonekawa, and
  Kurokawa}]{wang2020dnn}
Hanxin Wang, Daichi Amagata, Takuya Makeawa, Takahiro Hara, Niu Hao, Kei
  Yonekawa, and Mori Kurokawa. 2020.
\newblock A dnn-based cross-domain recommender system for alleviating
  cold-start problem in e-commerce.
\newblock \emph{IEEE Open Journal of the Industrial Electronics Society},
  1:194--206.

\bibitem[{Zhou and Agichtein(2020)}]{zhou2020rlirank}
Jianghong Zhou and Eugene Agichtein. 2020.
\newblock Rlirank: Learning to rank with reinforcement learning for dynamic
  search.
\newblock In \emph{Proceedings of The Web Conference 2020}, pages 2842--2848.

\bibitem[{Zhou et~al.(2020)Zhou, Agichtein, and
  Kallumadi}]{zhou2020diversifying}
Jianghong Zhou, Eugene Agichtein, and Surya Kallumadi. 2020.
\newblock Diversifying multi-aspect search results using simpson's diversity
  index.
\newblock In \emph{Proceedings of the 29th ACM International Conference on
  Information \& Knowledge Management}, pages 2345--2348.

\bibitem[{Zhou et~al.(2017)Zhou, Ni, and Rao}]{zhou2017block}
Jianghong Zhou, Jiangqun Ni, and Yuan Rao. 2017.
\newblock Block-based convolutional neural network for image forgery detection.
\newblock In \emph{Digital Forensics and Watermarking: 16th International
  Workshop, IWDW 2017, Magdeburg, Germany, August 23-25, 2017, Proceedings 16},
  pages 65--76. Springer.

\bibitem[{Zhou et~al.(2021)Zhou, Zahiri, Hughes, Al~Jadda, Kallumadi, and
  Agichtein}]{zhou2021biased}
Jianghong Zhou, Sayyed~M Zahiri, Simon Hughes, Khalifeh Al~Jadda, Surya
  Kallumadi, and Eugene Agichtein. 2021.
\newblock De-biased modeling of search click behavior with reinforcement
  learning.
\newblock In \emph{Proceedings of the 44th International ACM SIGIR Conference
  on Research and Development in Information Retrieval}, pages 1637--1641.

\bibitem[{Zhu et~al.(2019)Zhu, Mou, and Benyoucef}]{zhu2019exploring}
Wenlong Zhu, Jian Mou, and Morad Benyoucef. 2019.
\newblock Exploring purchase intention in cross-border e-commerce: A three
  stage model.
\newblock \emph{Journal of Retailing and Consumer Services}, 51:320--330.

\end{thebibliography}
\bibliographystyle{acl_natbib}

\end{document}